# Robot Composite Learning and the Nunchaku Flipping Challenge

Leidi Zhao, Yiwen Zhao, Siddharth Patil, Dylan Davies, Cong Wang, Lu Lu[1], Bo Ouyang[2]

*Abstract*— Advanced motor skills are essential for robots to physically coexist with humans. Much research on robot dynamics and control has achieved success on hyper robot motor capabilities, but mostly through heavily case-specific engineering. Meanwhile, in terms of robot acquiring skills in a ubiquitous manner, robot learning from human demonstration (LfD) has achieved great progress, but still has limitations handling dynamic skills and compound actions. In this paper, we present a composite learning scheme which goes beyond LfD and integrates robot learning from human definition, demonstration, and evaluation. The method tackles advanced motor skills that require dynamic time-critical maneuver, complex contact control, and handling partly soft partly rigid objects. We also introduce the "nunchaku flipping challenge", an extreme test that puts hard requirements to all these three aspects. Continued from our previous presentations, this paper introduces the latest update of the composite learning scheme and the physical success of the nunchaku flipping challenge.

## I. INTRODUCTION

We present a scheme of composite robot learning from human that empowers robots to acquire advanced manipulation skills that require

1) dynamic time-critical compound actions (as opposed to semi-static low-speed single-stroke actions),
2) contact-rich interaction between the robot and the manipulated objects exceeding that of firm grasping and requiring control of subtle bumping and sliding, and
3) handling of complex objects consisting of a combination of parts with different materials and rigidities (as opposed to single rigid or flexible bodies).

We also introduce the "nunchaku flipping challenge", an extreme test that includes hard requirements on all three elements listed above. Continued from our presentations in [1], [2], this paper introduces the latest updates of the proposed learning scheme and the experimental success of the nunchaku flipping challenge.

Advanced motor capabilities are without a doubt necessary for ubiquitous coexistence of robots and humans. Much research on robot dynamics and control does show success in realizing hyper robot motor capabilities. Representative work includes the running and hopping humanoid robot ASIMO by Honda Motor [3], the hyper balancing quadruped [4], biped [5], and wheeled [6] robots by Boston Dynamics, the high speed running cheetah robot by MIT [7], the dynamic vision guided baseball [8], regrasping [9], knotting [10], and pen spinning [11] robots by the University of Tokyo. Despite the application of adaptive and learning control, these works require extensive case-specific engineering that rely heavily on ad hoc models and control strategies, and lack scalability to ubiquitous applications.

Regarding the ubiquity of robot skill acquisition, a potential solution lies in robot reinforcement learning (RL) from trial and error as well as robot learning from human demonstration (LfD), which have become two hot topics in robotic research. First studied in the 1980s, LfD aims at providing intuitive programming measures for humans to pass skills to robots. [12] and more recently [13] give comprehensive surveys of LfD. Among the latest and most achieved, one well-known work is by the University of California, Berkeley, where a PR2 robot learned rope tying and cloth folding [14]. However, most LfD achievements so far are for semi-static decision-making actions instead of dynamic skills, partly due to the reliance on parameter-heavy computationally-intensive (for real-time evaluation) models such as deep neural networks (DNNs). In order to make the motion presentable to an audience, typical demonstration videos feature accelerated playback rates of up to ×50.

A few works in LfD do have achieved dynamic skills such as robots playing table tennis and flipping pancakes [15] as well as ball-paddling and ball-in-a-cup tricks [16], but with recorded human demonstration as the initial trajectories for the following robot reinforcement learning (RL) from (self) trial and error. The application of RL often features structurally parameterized control policies (e.g., [17], [18]) in the form of the combination of a few basis elements and can thus reduce the real-time computation load. The choice of the basis elements, however, are often quite case-specific. [19] gives a comprehensive survey of robot RL, which enables a robot to search for optimal control policy not from demonstrative data but from trial-and-error practice, with the goal of maximizing a certain reward function. Proper design of the reward function and the corresponding maximization strategy is another factor that is usually quite case-specific. The same authors of the survey (at Technische Universität Darmstadt) also achieved dynamic robot motor skills such as robots playing table tennis [20] and throwing darts [21] via applying RL and using motion primitives as basis elements. However, these works are mainly for stroke-based moves, and have not addressed compound actions.

Regarding these issues, we started studying a composite learning scheme, which showed success in a simulated nunchaku flipping test [1] and in an inverted pendulum swing-up experiment [2]. Since then, with the latest update of the scheme (Section II), we have achieved experimental success in the nunchaku flipping challenge (Sections III and IV).

[1] ECE, MIE, and ET Departments, New Jersey Institute of Technology, Newark, NJ 07102. {lz328,yz673,sp899,did3,wangcong,lulu}@njit.edu

[2] College of Electrical and Information Engineering, Hunan University, Changsha, China 410082. ouyangbo@hnu.edu.cn

## II. COMPOSITE SKILL LEARNING

So far, the majority of the robot learning from human research community has been focusing on the concept of robot learning from demonstrations (LfD). In many occasions, LfD has become a synonym of robot learning from human [13]. A few went further and explored techniques such as allowing robot learners to ask questions [22] and human mentors to give critiques [23] along with demonstration. Theories of human learning point out that effective learning needs more than observation of demonstrations [24]. In particular, explicit explanation of the underlying principals (e.g., [25], [26]) and testing with feedbacks (e.g., [27], [28], [29]) are necessary in effective teaching of complex skills. Expecting a learner to master new skills solely from observing demonstrations is analogous to learning from a silent teacher, which certainly could only achieve limited outcomes. This explains why the reinforcement learning (RL) assisted LfD shows effectiveness in learning dynamic motor skills - because the RL is in some sense a follow-up testing and feedback mechanism.

In regards to the limit of LfD, we propose a composite learning method that integrates robot learning from definition, demonstration, and evaluation.

---

**Composite Skill Learning**

1. The human mentor gives initial definition of the skill using a Petri net;
2. The human mentor demonstrates the skill for multiple times and self-evaluates the demonstrations;
3. The robot learner starts from the initial definition, use the demonstration data to learn the control policies for each transition and the judging conditions specified in the definition;
4. the robot also learns the evaluation criteria from the mentor's self-evaluated demonstration;
5. The robot tries out the skill and uses the learned criteria to conduct self-evaluation;
6. Additional human evaluations are optional and might help improve the fidelity of the evaluation criteria learned by the robot;
7. **if** *evaluated as failure* **then**
8.     The robot checks the scores of subprocedures, locates problematic spots, and modifies the initial definition by creating an adaptive Petri net;
9. Go to 5;
10. **else if** *evaluated as success* **then**
11.     The robot weights up the data from the successful trials so as to improve the learned control policies and conditions;
12.     After reaching a stable performance above a certain successful rate, the skill is considered learned.

---

### A. Adaptive Learning from Definition

We use Petri nets (PN) to define compound skills that includes multiple subprocedures. The places in the PN consist of the state variables of the robot, such as posture, joint velocities, and torques. The states and transitions in a PN represent the subprocedures and the corresponding motion actions respectively. Each transition in the PN features a relatively coherent motion pattern and can be realized using a single motion/force control policy. Petri nets are abstract enough to be composed intuitively by humans, while sufficiently symbolic for machines to parse. Despite being widely used in robotics (e.g., [30], [31]), Petri nets have yet not been used to teach robots dynamic skills.

Due to possible improper human description and very often the physical difference between the human mentor and the robot learner, modification of the initial definition is necessary. Starting from an initial definition provided by the human mentor, we use adaptive measures to enable autonomous correction of the initial definition. Instead of a standard 4-tuple Petri net $PN = (P, T, A, M_0)$, we introduce a 6-tuple adaptive Petri net $APN = (P, T, A, M_0, \Lambda, C)$, where $P$ is the set of places, $T$ is the set of transitions, $A$ is the incident matrix that defines the relationship among places and transitions, $M_0$ is the initial marking, $\Lambda$ is a set of firing probabilities of transitions, and $C$ is a set of firing conditions.

An APN allows the robot learner to revise the initial definition through learning from evaluation (Section II-C). By adjusting the $P$ set, $T$ set and $A$ matrix, places and transitions can be added or dropped from the initial definition. Our previous paper [2] presented an inverted pendulum test, in which a transition is added by the learning agent to recover from a wrong state. In addition, adjustment of the firing probability set $\Lambda$ and the condition set $C$ changes the learned skill towards more suitable to the mechanical characteristics of the robot. Section IV gives an example.

The state equation of the Petri net is

$$M' = M + A\mu \tag{1}$$

where the $M$ is the previous marking, $M'$ is the marking after a transition fires. $\mu$ is a column vector indicating whether the transitions fire with its boolean elements. It is controlled by the set of firing probability $\Lambda$ and the set of conditions $C$

$$\mu = \begin{bmatrix} d_0 p_0 \\ d_1 p_1 \\ d_2 p_2 \\ \vdots \end{bmatrix} \tag{2}$$

where $d_i$ is a boolean decision value indicating if the firing condition $c_i \in C$ of the $i$th transition is satisfied. $p_i$ is a boolean random value that follows Bernoulli distribution $\Pr(p_i = 1) = \lambda_i$, where $\lambda_i \in \Lambda$ defines the firing probability of the $i$th transition.

Starting from the initial $C$ and $\Lambda$ assigned by the human mentor, the robot carries out modification through trying out the skill. When a problematic transition is identified, its firing probability $\lambda_i$ is updated to

$$\lambda_i^* = \kappa \lambda_i \tag{3}$$

where $\kappa < 1$. Once $\lambda_i$ drops below a certain level, the corresponding firing condition $c_i$ will be updated to

$$c_i^+ = \begin{bmatrix} w_1 & \cdots & w_k \end{bmatrix} \begin{bmatrix} s_i^1 & \cdots & s_i^k \end{bmatrix}^\mathsf{T} \quad (4)$$

where $w_j$ is a weight parameter derived from the evaluation of the $j$th trial. $s_i^j$ is the recorded state when firing the $i$th transition at $j$th trial. The firing probability resets when the corresponding condition is updated.

### B. Learning from Demonstration with Data Conditioning

The Petri net definition divides a compound skill in a way that each transition has a relatively coherent motion pattern and can be governed by a single control policy regressed from the human demonstration data. To avoid case-specific engineering of model-based control, we use nonparametric regression methods. Nowadays, more and more research involving nonparametric learning use deep neural networks (DNNs) with convolutional or spiking modules, taking the advantage that a large amount of training parameters (e.g., 18 million parameters in [32]) benefits the approximation of complicated state-control mappings. The price, however, is the difficulty of executing the learned control policy in real-time for dynamic actions. [33] combined DNN with parametrized policy search and obtained a model of relatively smaller scale with around 92 000 parameters. The reduced size, however, still only allows a control rate of 20 Hz, which is difficult for dynamic actions that usually require a control rate at several hundreds to over a thousand Hz.

Instead of counting on standalone LfD with a huge amount of parameters, we seek breakthrough from the power of composite learning and turn to the more computationally efficient Gaussian Process Regression (GPR), aiming at realizing a high control rate with regular control systems. GPR has a strong history in learning control of robots. One pioneering work is presented in [34]. In our work, the regression learns a mapping from the system state $x$ to the control $u$. The learned mapping is used as the motion control policy to realize a specific transition in the PN definition. Consider a set of data $\{(x_i, u_i) : i = 1, 2, \ldots, n\}$ from human demonstrations. GPR assumes that the mapping $u = u(x)$ follows a multi-variable joint distribution with certain statistical characteristics. We apply the squared exponential kernel function in the state space

$$k(x_i, x_j | \theta) = \sigma^2 \exp\left(-\frac{1}{l^2}(x_i - x_j)^\mathsf{T}(x_i - x_j)\right) \quad (5)$$

where $\theta = \{\sigma, l\}$ includes the so-called hyperparameters to be trained, with $\sigma$ being a covariance scaling factor and $l$ being a distance scaling factor. Because the whole skill is divided into multiple subprocedures that each has a relatively simple motion pattern, there is no need to use advanced kernels (e.g., [35] Section 5.4.3 ), which lead to demanding and case-specific parameter training. The covariance matrix of the data is

$$K = \begin{bmatrix} k(x_1, x_1) & \cdots & k(x_1, x_n) \\ \vdots & \ddots & \vdots \\ k(x_n, x_1) & \cdots & k(x_n, x_n) \end{bmatrix} \quad (6)$$

and the covariance matrix relating the queried state $x_*$ to the data is

$$K_* = \begin{bmatrix} k(x_*, x_1) & \cdots & k(x_*, x_n) \end{bmatrix} \quad (7)$$

The control $u_*$ for the queried state $x_*$ can be inferred using the conditional expectation

$$\mathbb{E}[u_*] = K_* K^{-1} U_t \quad (8)$$

where $U_t$ is the stack of the controls in the training data. Note that the computation load of $u_*$ very much depends on the sizes of $K$ and $K_*$, which in turn depend on the size of the data set. In order to achieve high computing efficiency for real-time control as well as improve the fidelity of the regressed mapping, we have developed a data conditioning method [36] using rank-revealing QR (RRQR) factorization. The RRQR factorization of the stack $S_t$ of the states from the data set is in the form of

$$S_t \Pi = Q \begin{pmatrix} R_{11} & R_{12} \\ \mathbf{0} & R_{22} \end{pmatrix} \quad (9)$$

where $\Pi$ is a permutation matrix, $Q$ is orthogonal, and $R_{11} \in \mathbb{R}^{m \times m}$ is well conditioned. The columns in $S_t$ identified by the first $m$ columns of $\Pi$ form a well conditioned subset. Various algorithms are available to compute an RRQR factorization, providing different lower bounds of $R_{11}$'s smallest singular value, with a general form of

$$\sigma_m(R_{11}) \geqslant \frac{\sigma_m(S_t)}{l(m, n)} \quad (10)$$

where $l(m, n)$ is bounded by a polynomial of $m$ and $n$ [37]. The subset selected from the raw data stack features improved condition number and leads to more reliable regression fidelity, while takes only a fraction of the original computation.

### C. Learning from Evaluation

After acquiring the skill definition and regressing control policies from human demonstration, we use evaluations to tune the learned skill. The human mentor and the robot learner often have nontrivial physical difference, and the skills learned right off the human demonstration are often not optimal or even less feasible to the robot learner. Learning from evaluation handles this problem. When the robot tries to carry out a learned skill, both the success of the whole skill and the performance of each subprocedure will be evaluated. In order to avoid case-specific engineering, the scoring formulae are not explicitly specified by the human mentor. Instead, the human mentor labels his/her demonstration with success/failure flags and performance scores. The learning agent learns the scoring criteria from the labeled data and use it to self-evaluate the robot's practices, which always have variations due to the dynamic and compound nature of the skills. The evaluation result is in turn used to refine the learned skills by taking in the data from more successful practices, while the data from lower scored demonstrations are less weighted. Examples are discussed at the end of Section IV.

## III. HARDWARE PREPARATION

### A. A robot arm and real-time control system

We use a 6-DOF AUBO i5 6-joint robot arm. It features a modular design similar to the popular UR series by Universal Robots, while provides a much simpler open control interface that allows the users to fully access real-time position/velocity/torque control. The control deployment (Fig. 1) is based on a target computer directly connected to the robot arm via a Controller Area Network (CAN bus) cable. Other than a National Instruments PCI-CAN interface, no additional interfacing hardware is used. MATLAB/Simulink is used to implement sensing, control, and safety algorithms. The MATLAB Vehicle Network Toolbox is used to facilitate the CAN communication protocol. The sampling rate of the control system is 1 kHz.

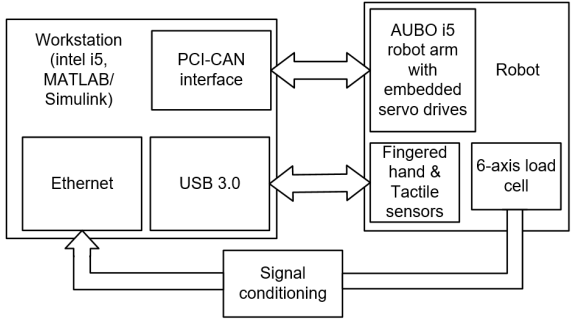

Fig. 1. Control deployment of the test setup

### B. A bionic robot hand

In order to facilitate advanced manipulation skills involving finger actions, we developed a bionic robot hand with haptic sensors. The hand features a bionic five-finger design and tendon-driven actuation. The majority of the hand is 3D-printed, including the palm and finger segments in PET, finger joints in the rubbery TPU (for auto extension), and a motor pack in stainless steel. TakkTile sensors developed by Righthand Robotics are used as haptic sensors. They are built up on the NXP MPL115A2 MEMS barometer by casting a rubber interface [38]. In addition, an ATI 6-axis load cell is installed at the wrist to perceive the centrifugal force brought by the motion of any payload manipulated by the hand.

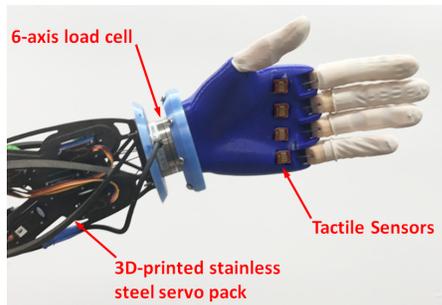

Fig. 2. A fingered robot hand with haptic sensors

### C. Motion capture systems

Motion capture is used in learning from demonstration. Accurately capturing the necessary details of a highly dynamic skill is a challenge. It is also important to provide an intuitive interface for efficient teaching. In order to satisfy these criteria, we have experimented several options.

The Microsoft Kinect seems to be a first choice because it offers real-time bare-hand motion capture, and has been recognized as a top product among commercial camera and image processing-based systems. Not requiring markers or hand-held gadgets makes it very intuitive for the mentor to demonstrate a skill. However, such camera-based systems suffer from limited sampling rate (usually up to 30 frames per second) and considerable delay caused by image processing (usually takes up to an entire sampling period), plus low accuracy as reported in [39], which in turn make the velocity estimation difficult. We tried to compensate these problems, otherwise known as visual sensing dynamics using a predictive filtering technique [40]. First, the position signal $s(t)$ being sensed is decomposed using it's Taylor expansion with respect to time

$$s(t_0 + t) = \sum_{c=0}^{r} s^{(c)}(t_0) \frac{t^c}{c!} + s^{(r+1)}(t_0 + \xi) \frac{t^{r+1}}{(r+1)!} \quad (11)$$

where $r$ is the order of the expansion, and $\xi \in (0, t)$. The expansion can be written into a state space model:

$$x(i+1) = \begin{bmatrix} 1 & T & \cdots & \frac{T^r}{r!} \\ & 1 & \cdots & \frac{T^{r-1}}{(r-1)!} \\ & & \ddots & \vdots \\ & & & 1 \end{bmatrix} x(i) + \begin{bmatrix} \frac{T^{r+1}}{(r+1)!} \\ \frac{T^r}{r!} \\ \vdots \\ T \end{bmatrix} u(i) \quad (12)$$

with an output

$$y(j) = \begin{bmatrix} 1 & 0 & \cdots & 0 \end{bmatrix} x(j - L) + v(j) \quad (13)$$

where the state vector $x = \begin{bmatrix} s & s' & \cdots & s^{(r)} \end{bmatrix}^\mathsf{T}$ contains $s$ and its derivatives. $i$ is the time step index. $T$ is the algorithm sampling time, which is much shorter than the camera sampling time. $u$ comes from the residual term in Eq. (11). It is treated as an unknown input, and handled using an *equivalent noise approach* [41]. $y$ is the position identified by image processing. $v$ is the artifacts and rounding noise. $j = N, 2N, 3N, \ldots$ is the index of the camera sampling actions. $L$ is the delay caused by image processing.

A dual-rate adaptive Kalman filter is then be applied to Eq. (12) and (13) to compensate for the delay and recover the information between sampling actions. Despite reported success of this type of compensation techniques [40], we found that it still requires the camera to sample at least 15 times faster than the desired bandwidth of the motions being sensed. For the highly dynamic maneuvers targeted in our work, such a limit excludes the use of any commercial camera and image processing-based motion capture systems.

Another type of non-contact motion capture systems is the ones using active infrared makers and infrared sensors. Consumer level products of such type include the Nintendo

Wii and Sony PS Move, which unfortunately are of very limited accuracy [42]. Meanwhile, the high-end options such as the one introduced in [43] and its commercial peers (e.g., PhaseSpace and NDI Optotrak), although are capable of obtaining very high quality measurement, are much beyond the space and cost considerations in our long term goal of making the technology available to everyday life.

A balanced choice between cost and capability is a mechanical motion capture system in the form of a passive multi-bar mechanism equipped with rotation sensors at the joints. Compared to the previous two options, such systems, either the commercial ones such as Geomagic Touch or a customized design (Fig. 3 left) provide both an feasible cost and sufficient capability to our application. Although the usage is not as intuitive as non-contact motion capture systems, the additional difficulty is acceptable. In addition, a sensing glove with Flex sensors is used to capture the motion of the fingers. The glove is also equipped with vibrating motors to provide tactile feedback to the user (Fig. 3 right).

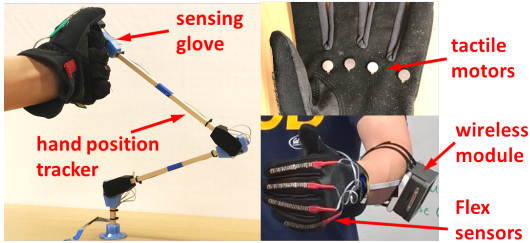

Fig. 3. Motion capture gadgets used in the tests: Left: a mechanical motion capture device. Right: a wireless sensing glove.

## IV. THE NUNCHAKU FLIPPING CHALLENGE

We introduce *the nunchaku flipping challenge* to test the proposed learning scheme. Nunchaku is a traditional Okinawan martial arts weapon widely known due to its depiction in film and pop culture. It consists of two rigid sticks connected by a chain or rope. Among the numerous tricks of using nunchaku, the flipping trick as shown in Fig. 4 is one that puts hard challenges to all three elements we consider in advanced manipulation skills, i.e., dynamic maneuver, hand-object contact control, and handling partly soft partly rigid objects. The trick includes three subprocedures: swing-up (1–3 in Fig. 4), chain rolling (4–6), and regrasping (7, 8).

With the composite learning scheme, the nunchaku flipping trick is first described by a human mentor using an initial Petri net definition. As shown in Fig. 5, $P_0$ is the initial state, in which the robot hand holds one of the sticks and is in no motion. When the start-of-motion transition $t_0$ fires, the robot begins the swing-up procedure. The swinging $t_1$, the stop motion $t_6$, and the hand-releasing action $t_2$ fire based on the probability and the judging conditions in $P_1$. If the running time goes beyond a threshold, the action stops by firing the stop motion $t_6$. If the sensor reading is below a certain level, the swing $t_1$ fires and the amplitude of the swing is increased. After the sensor reading exceeds the threshold,

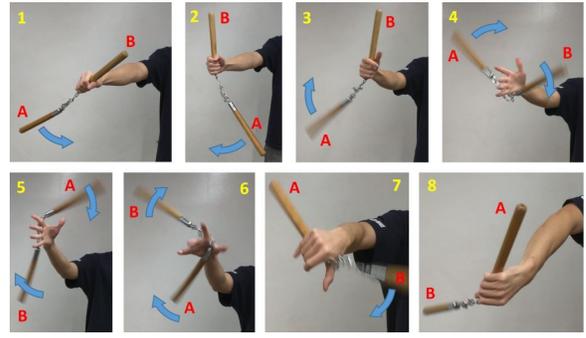

Fig. 4. The nunchaku flipping trick

$t_2$ has a possibility to fire. Similar to the swing-up procedure, when the hand-releasing action $t_2$ successfully fires, the robot goes on to chain-rolling. The back palm contact control $t_3$, the stop motion $t_7$, and the regrasping action $t_4$ fire based on the probability and the judging conditions in $P_2$. The robot regrasps by firing $t_4$. If the regrasping is successful according to the condition in $P_3$, stop motion $t_5$ fires and leads to the final success $P_{F(\text{success})}$. Otherwise, stop motion $t_8$ fires and leads to the final failure $P_{F(\text{fail})}$. The possibilities and conditions could change during the learning process.

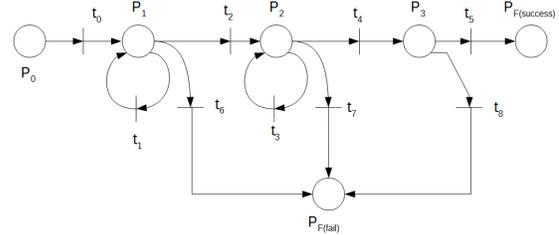

Fig. 5. Initial Petri net definition of the nunchaku flipping trick

The bionic robot hand described in Section III-B is installed on the robot arm to resemble human hand maneuvering. In order to keep a reasonable fidelity to human sensory control, no sensor is installed on the nunchaku. The motion of the sticks and the chain is perceived by the haptic sensors and 6-axis load cell in the robot hand. In addition, without explicit inference of the position, attitude, and layout of the sticks and the chain, the sensor readings are directly mapped to the motor controls of the fingers and the arm joints by the learning algorithm. Such an end-to-end learning scheme has earned increasing preference recently and is believed to be a good approximation of human neural response (e.g., [33]).

Multiple demonstrations of the nunchaku flipping trick are performed by a human mentor and recorded by the motion capture systems. The human mentor labels if each demonstration is a success and scores the performance. The control policies of the transitions and the judging conditions are learned from successful demonstrations weighted by their scores. The grading criteria for robot self-evaluation are learned from both successful and failed demonstrations. Starting with the initial definition, the robot conducts multiple trials. After each trial, the robot grades its own performance using the learned criteria. The final score and the score of every transition are given to determine if the trial is a success and which part in the definition should be adjusted.

Such adjustment is important because of the physical differences between the human mentor and the robot learner. This is especially true for the ending part of the swing-up which requires certain vertical speed to enter chain-rolling. The human mentor tends to use a sudden jerk-up to realize this part. Despite being a small move, this action is at the border of the robot's mechanical limit. As a result, the learning agent avoids learning from demonstrations featuring this move because of a low success rate during the trial runs, while weights up the data form demonstrations with a more back-and-forth type swing-up, which achieves much more successes in the trial run evaluations. Similar situation applies to the condition of switching from swing-up to chain rolling. The initial switching condition learned from human demonstrations is not the optimal for the robot, which can be adjusted through learning from evaluation during the trial runs.

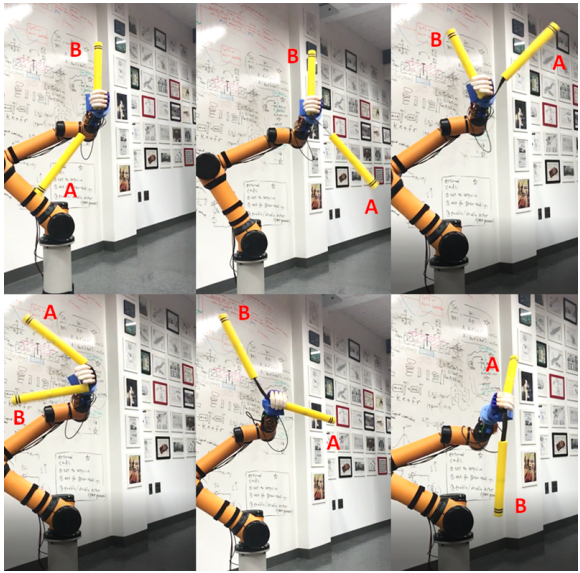

Fig. 6. Nunchaku flipping learned by a robot (video uploaded to PaperPlaza)

## V. Conclusions

In regard to the difficulties of robot learning from demonstration on tackling dynamic skills and compound actions, this paper introduces a composite robot learning scheme which integrates adaptive learning from definition, nonparametric learning from demonstration with data conditioning, and learning from evaluation. The method tackles advanced motor skills that require dynamic time-critical maneuver, complex contact control, and handling partly soft partly rigid objects. We also introduce the "nunchaku flipping challenge", an extreme test that puts hard requirements to all these three aspects. Details of the hardware preparation and control system deployment of a physical test are explained. The proposed robot learning scheme shows promising performance in the challenge.

Future work will focus on introducing a fusing method to merge the learning from multiple human mentors, as well as testing the composite learning method with additional challenges, including more advanced tricks of nunchaku usage and skills of handling other complex objects such as yo-yo and kendama. Hardware-wise, vibration suppression of the fingers during the highly dynamic motion is an issue.